\documentclass[letterpaper]{article} 
\usepackage{aaai2027} 
\usepackage[hyphens]{url} 
\usepackage{graphicx} 
\usepackage{natbib} 
\usepackage{caption} 
\usepackage{amsmath}
\usepackage{amssymb}
\usepackage{booktabs}
\usepackage{multirow}

\title{PhotoHOI: Synthesizing 3D Hand-Object Interactions from a Single RGB Photograph}
\author{
    Zhenhao Zhang\textsuperscript{\rm 1},
    Jiajun Zhang\textsuperscript{\rm 1},
    Wei MIN\textsuperscript{\rm 2},
    Yebin Liu\textsuperscript{\rm 1}\corresponding
}
\affiliations{
    \textsuperscript{\rm 1}Tsinghua University
    \textsuperscript{\rm 2}Shadow AI\\
}

\begin{document}

\maketitle

\begin{abstract}
Hand-object interaction (HOI) is a fundamental human behavior
with broad applications in AR/VR, digital humans, and embodied
interaction.
Existing methods typically require predefined object geometry,
object trajectories, or task-specific conditions, limiting their use with natural real-world inputs.
To address this, we study a more practical problem of synthesizing
3D hand-object interaction sequences from a single RGB
photograph and an open-vocabulary language instruction, and
introduce PhotoHOI.
PhotoHOI first uses a vision-language model to parse the input
image and instruction into a structured task specification,
including the interaction object, target region, and spatial
relation.
It then recovers a compact task-relevant 3D scene and plans a
smooth collision-aware object trajectory based on the recovered
object states, support relations, and surrounding scene geometry.
To synthesize hand motion that generalizes to real-world
photographs and unseen objects, it learns transferable
task-conditioned contact and contact-conditioned grasp priors
from large-scale affordance and HOI data.
The grasp is further refined in a learned latent space, constraining the optimization to a plausible hand-pose manifold.
Experiments on GRAB and H2O demonstrate improved contact quality
and reduced penetration over representative baselines.
Results on real-world photographs further demonstrate higher task
success and scene consistency, together with generalization to
unseen objects and open-vocabulary instructions.
\end{abstract}

\section{Introduction}

\begin{figure*}[t]
  \centering
  \includegraphics[width=\textwidth]{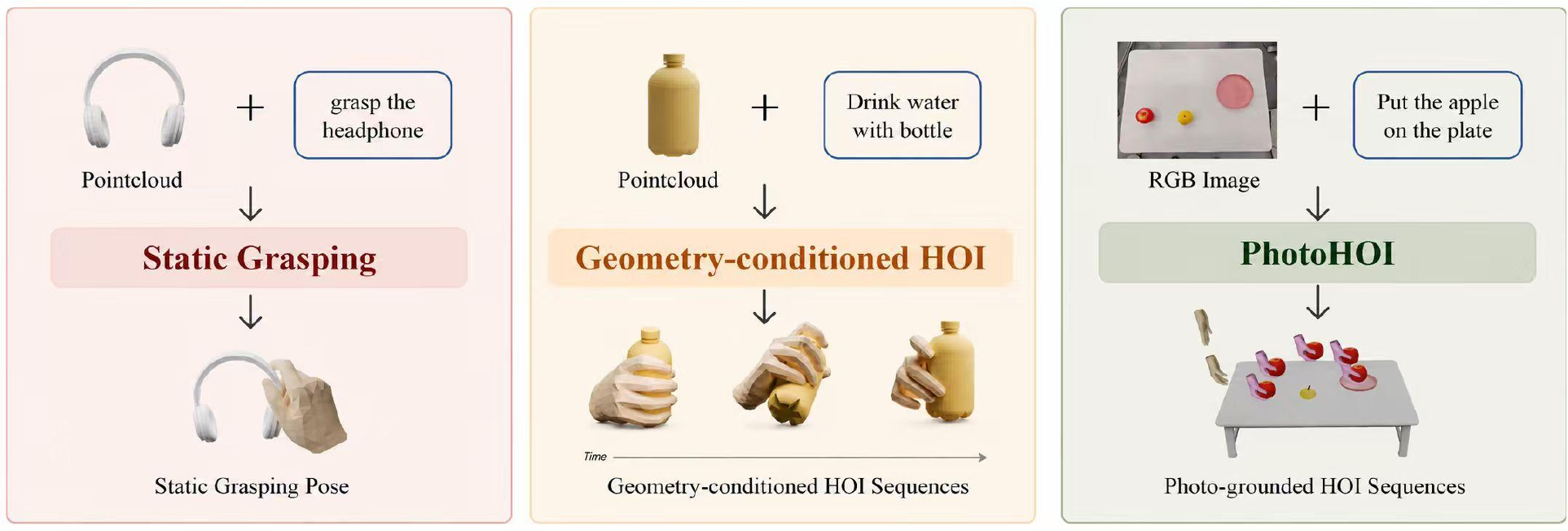}
  \caption{
  Overview: Unlike existing methods that rely on known object
  geometry or given trajectories, PhotoHOI starts from a single
  RGB photograph and an open-vocabulary instruction to generate
  a 3D hand-object interaction sequence grounded in the recovered
  task-relevant scene.
  }
  \label{overview}
\end{figure*}

Hand-object interaction (HOI) is a common form of human
interaction in daily life and has broad applications in AR/VR
\cite{holl2018efficient,mangalam2024enhancing,tang2023cafi},
digital humans
\cite{xu2025interact,chen2025interactavatar},
and embodied manipulation
\cite{wang2022dexgraspnet,xu2023unidexgrasp,
zhu2025evolvinggrasp,fang2025anydexgrasp}.
However, many existing HOI synthesis methods require
predefined object geometry, object trajectories, or other
task-specific conditions that must be prepared separately and
cannot be obtained directly from a single photograph.
A practical system should instead operate from natural user
inputs, such as an RGB photograph and an open-vocabulary
instruction, and generate a 3D interaction sequence grounded
in the recovered task-relevant scene.
As illustrated in Fig.~\ref{overview}, this setting reduces
the reliance on manually prepared object geometry and
trajectories.

Recent advances in HOI synthesis have made significant
progress in generating plausible hand-object interactions.
Early methods
\cite{liu2023contactgen,zhang2024graspxl,
zhong2025dexgrasp}
mainly focus on synthesizing static grasps for a given object.
More recent works extend this setting to dynamic interaction
sequence generation by using text descriptions
\cite{cha2024text2hoi,christen2024diffh2o},
contact maps
\cite{zhang2025bimart,han2025touch,
yan2025dexterous,yu2025dynamic},
or object trajectories
\cite{zhang2026handx,zhang2025manidext,chen2025fbi}
as conditions.
Despite these advances, most existing methods still require
object geometry or motion-related conditions to be specified
in advance and therefore cannot directly synthesize interactions
from natural image-language inputs.

To bridge this gap, we present PhotoHOI, a framework for
synthesizing 3D hand-object interactions from minimal
user-provided inputs.
Given only a single RGB photograph and an open-vocabulary
language instruction, PhotoHOI generates a 3D hand-object
interaction sequence grounded in the recovered task-relevant
scene.
We devise a two-stage approach that first performs task
parsing, task-relevant scene recovery, and geometry-guided
object-motion planning, and then synthesizes the corresponding
hand motion.
The central challenge is to infer object geometry, initial
object states, target relations, and interaction conditions that
are explicitly provided to many existing HOI synthesis methods
but are unavailable in our image-language input setting.

Natural-language instructions generally describe interaction
goals at a semantic level, whereas downstream 3D recovery
and motion synthesis require explicit objects, spatial relations,
and geometric states.
To bridge this discrepancy, PhotoHOI integrates VLM-based
task parsing, task-relevant 3D scene recovery, and object-motion
planning.
Given an input photograph and language instruction, a
vision-language model extracts structured task cues, including
the interaction object, target object or region, action type, and
spatial relation.
Guided by these cues, PhotoHOI segments and reconstructs
the objects and support structures relevant to the task.
The recovered object geometries and poses are refined using
support information and are then used to determine the initial
and target object states.
Based on these states and the parsed spatial relation, PhotoHOI
plans a smooth collision-aware object trajectory for subsequent
hand-motion synthesis.

Recovering an object trajectory alone is insufficient for
synthesizing plausible hand motion, especially when the
reconstructed objects differ from those observed in the HOI
training data.
Generalizing hand synthesis to such objects requires
interaction priors that capture transferable functional contact
and grasp patterns rather than object-specific configurations.
To this end, we first learn a task-conditioned contact prior
from a large-scale collection of affordance and HOI data.
The contact prior predicts functional surface regions that guide
hand placement on both seen and unseen object geometries.
We further learn a contact-conditioned grasp prior that maps
the object geometry and predicted contact regions into a
compact latent distribution of plausible hand configurations.
During inference, instead of directly optimizing
high-dimensional hand articulations, PhotoHOI
refines the grasp through the learned latent variables together
with the global wrist pose.
This constrains the optimization to a plausible hand-pose
manifold while aligning the hand with the predicted contact
regions and reducing hand-object interpenetration.
After obtaining the refined grasp, PhotoHOI generates an
approach motion and coordinates the hand with the planned
object trajectory during manipulation.

Experiments on standard HOI benchmarks and real-world photograph input demonstrate that PhotoHOI generates physically plausible and semantically consistent 3D hand-object interaction sequences from minimal image-language inputs. The results suggest a practical step toward deploying 3D HOI synthesis in real-world scenarios. 
Our contributions are summarized as follows:

\begin{itemize}

    \item
    We introduce PhotoHOI, a framework that synthesizes
    3D hand-object interaction sequences from a single RGB
    photograph and an open-vocabulary instruction, reducing
    the need for manually prepared object geometry and
    trajectories.

    \item
    We develop an image-to-motion pipeline that parses
    task-relevant objects and spatial relations, recovers their
    geometry and poses, and plans a smooth collision-aware
    object trajectory based on the recovered scene.

    \item
    We develop transferable interaction priors for hand
    synthesis by combining a task-conditioned contact prior
    with a contact-conditioned grasp prior.
    By refining the grasp in the learned latent space, PhotoHOI
    maintains plausible hand articulation while satisfying
    contact and penetration objectives on reconstructed objects.

\end{itemize}
\section{Related Work}

\subsection{Hand-object Interaction Synthesis}

\textcolor{black}{
3D hand-object interaction synthesis has progressed from static grasp generation to dynamic interaction sequence generation.
Text2HOI\cite{cha2024text2hoi} decomposes text-conditioned HOI synthesis into contact and motion generation, while DiffH2O\cite{christen2024diffh2o} generates one- or two-handed interactions from language and object geometry.
HOIGPT\cite{huang2025hoigpt} represents language and HOI sequences in a unified token space, OpenHOI\cite{zhang2025openhoi} supports free-form instructions and unseen objects, and TOUCH\cite{han2025touch} leverages large-scale videos to improve interaction diversity.
Despite these advances, most existing methods still assume that object geometry or motion-related conditions are provided.
In contrast, PhotoHOI derives task-relevant object geometry and object-motion conditions from a single RGB photograph and an open-vocabulary instruction before synthesizing the corresponding hand motion.
}

\subsection{Vision Language Models for Digital Human}

\textcolor{black}{
Vision-language models and large language models provide semantic understanding and task reasoning capabilities for interaction generation.
SemGrasp\cite{li2024semgrasp} introduces language semantics into grasp generation, while MotionGPT\cite{jiang2023motiongpt} and MotionChain\cite{jiang2024motionchain} model language-conditioned human motion.
HOIGPT\cite{huang2025hoigpt} applies language modeling to 3D HOI sequences, and VLM-RMD\cite{deng2025human} uses VLM reasoning to guide human-object motion policies.
Different from these works, PhotoHOI uses a VLM as a task parser that converts an image-instruction pair into structured objects, actions, and spatial relations for subsequent 3D recovery and motion synthesis.
}

\subsection{Hand-object Interaction Video}

\textcolor{black}{
Videos provide large-scale observations for learning hand-object interaction dynamics.
SViMo\cite{dang2025svimo} and HarmoHOI\cite{dang2026harmohoiharmonizingappearance3d} jointly model HOI videos and 3D motion, while PlayerOne\cite{tu2026playerone} and EgoSim\cite{hao2026egosim} focus on egocentric world simulation.
Other methods use human activity videos to learn interaction representations or robotic manipulation policies
\cite{li2025scalable,luo2025being,zhang2026unihm,li2026gazevla}.
These approaches mainly target video generation, world modeling, or robotic action learning, whereas PhotoHOI synthesizes explicit 3D hand and object motion sequences from a single RGB photograph and a language instruction.
}

\begin{figure*}[t]
  \centering
  \includegraphics[width=\textwidth]{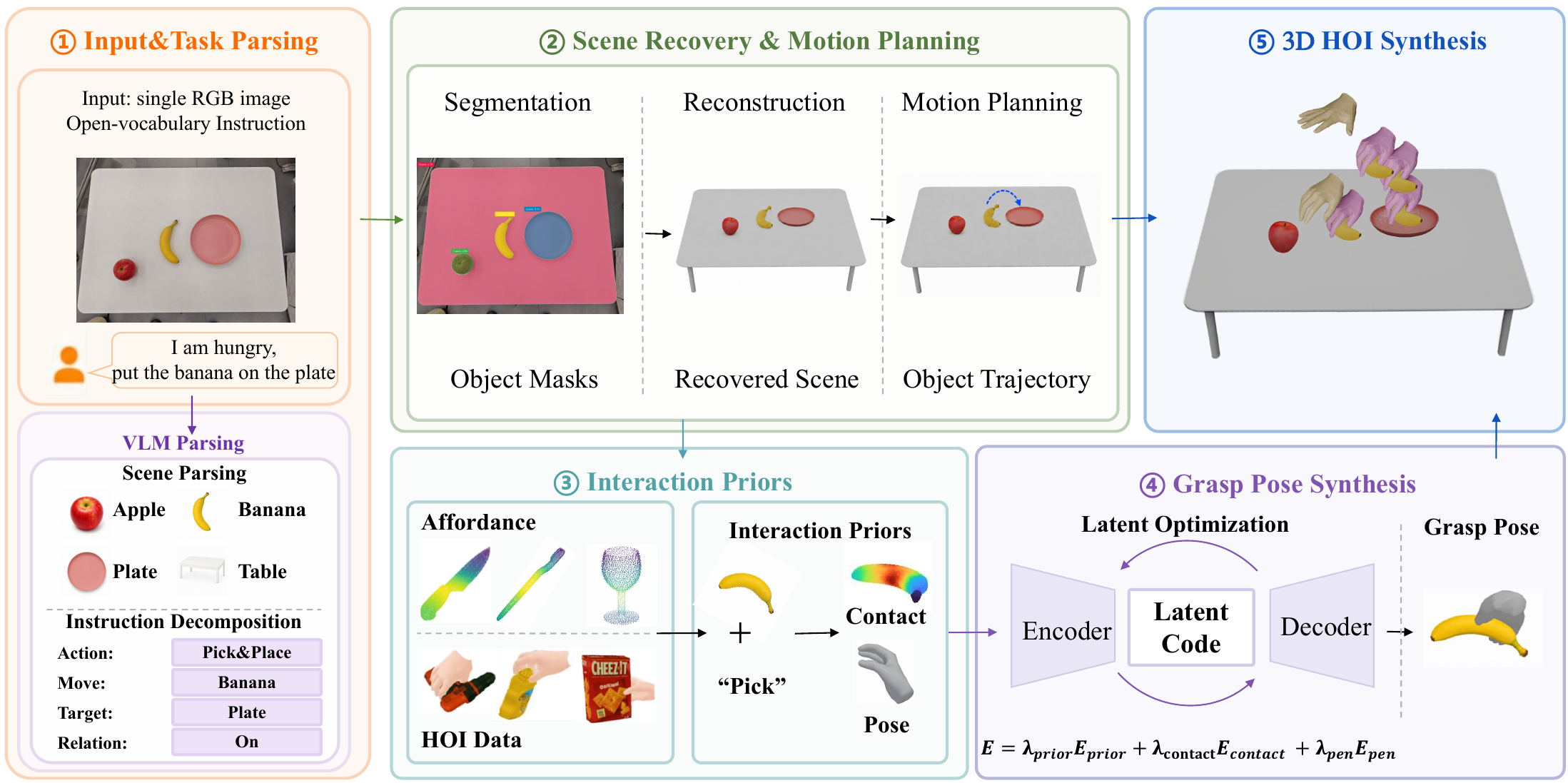}
  \caption{Given a single RGB photograph and an instruction, \textcolor{black}{PhotoHOI} parses the task, recovers the task-relevant 3D scene, and plans a collision-aware object trajectory. It then predicts interaction priors and refines the grasp in a learned latent space. The resulting hand motion and object trajectory form a scene-grounded 3D hand-object interaction sequence.
  }
  \label{fig:pipeline}
\end{figure*}

\section{Method}

\subsection{Problem Formulation}
Given a real-world RGB photograph $I$ and an open-vocabulary language instruction $g$, PhotoHOI aims to recover a task-relevant 3D scene representation $\mathcal{S}$ from the image and generate a 3D hand-object interaction sequence $\mathcal{Y}$ grounded in the recovered objects and their spatial relations.
The overall pipeline is illustrated in Fig.~\ref{fig:pipeline}. We denote:

\begin{equation}
\begin{gathered}
(\mathcal{Y},\mathcal{S})=\mathrm{PhotoHOI}(I,g), \\
\mathcal{Y}=\{H_t,T_t^{\mathrm{obj}}\}_{t=1}^{T},\qquad
\mathcal{S}=\{(G_i,T_i)\}_{i=1}^{N}.
\end{gathered}
\label{eq:photohoi_output}
\end{equation}
where $H_t$ denotes the hand pose and articulation state at frame $t$, and $T_t^{\mathrm{obj}}$ denotes the 6DoF pose of the interaction object at frame $t$ in the recovered scene coordinate system.
$G_i$ and $T_i$ denote the local geometry of the $i$-th recovered object and its similarity transform in the reconstructed scene $\mathcal{S}$, respectively.
\textcolor{black}{
Unlike methods that assume predefined object geometry or object-motion conditions, our setting requires the system to infer task-relevant object geometry and poses, construct an object trajectory that accounts for the recovered scene geometry, and synthesize the corresponding hand motion from image-language inputs.
}

\subsection{VLM-based Task Parsing}

Given the input image $I$ and language instruction $g$, the first step is to ground the open-vocabulary task in the visual scene.
This is non-trivial because human instructions usually describe tasks at a high semantic level, whereas downstream 3D recovery and motion planning require structured cues, including task-relevant objects, target regions, and spatial relations.
\textcolor{black}{
We therefore use a VLM~\cite{team2023gemini} as a semantic task parser that converts the image-instruction pair into a compact structured specification for subsequent scene recovery and motion synthesis.
}

Specifically, the VLM outputs a set of task-relevant objects $\mathcal{O}_{\mathrm{task}}$ and a structured task specification $\tau$:
\begin{equation}
\begin{gathered}
(\mathcal{O}_{\mathrm{task}},\tau)=\mathrm{VLM}(I,g), \\
\tau=(a,o_{\mathrm{int}},o_{\mathrm{goal}},r).
\end{gathered}
\label{eq:vlm_parse}
\end{equation}

Here, $\mathcal{O}_{\mathrm{task}}$ contains the task-relevant objects or regions that need to be localized and processed for the current task.
The task specification $\tau$ consists of the action type $a$, the interaction object $o_{\mathrm{int}}$, the target object or region $o_{\mathrm{goal}}$, and the spatial relation $r$.
For example, given ``I am hungry, place the banana on the plate,'' the VLM parses \textit{place} as $a$, the \textit{banana} as $o_{\mathrm{int}}$, the \textit{plate} as $o_{\mathrm{goal}}$, and \textit{on} as $r$.

In this way, VLM parsing converts free-form image-language inputs into structured cues for downstream modules:
$\mathcal{O}_{\mathrm{task}}$ guides object localization and scene recovery;
$o_{\mathrm{int}}$, $o_{\mathrm{goal}}$, and $r$ determine the object-motion planning problem;
and $a$ conditions the contact prior.
\textcolor{black}{
The VLM provides semantic task cues rather than metric 3D geometry; object geometry, poses, scales, support relations, and obstacle geometry are estimated in the subsequent scene recovery stage.
}

\subsection{Scene Recovery}

Given the task-relevant object set $\mathcal{O}_{\mathrm{task}}$ parsed by the VLM, scene recovery lifts the relevant image observations into a 3D representation.
We build on visual and 3D foundation models for open-vocabulary segmentation and single-image reconstruction to obtain object geometry and initial poses, followed by a lightweight support-aware refinement.
This stage provides the geometric and spatial context required for subsequent trajectory planning and hand synthesis.

\paragraph{Open-vocabulary Object Segmentation.}

We first localize the task-relevant objects in the image and obtain their instance masks using an open-vocabulary detection and segmentation foundation model~\cite{ren2024grounded}:
\begin{equation}
\mathcal{M}
=
\Phi_{\mathrm{seg}}(I,\mathcal{O}_{\mathrm{task}})
=
\{M_i\}_{i=1}^{N}.
\label{eq:object_segmentation}
\end{equation}
Here, $M_i$ denotes the binary instance mask of the $i$-th object.
\textcolor{black}{
The segmentation step focuses on the regions required by the interaction task, including the interaction object, the target object or region, support structures, and nearby objects that may obstruct the planned motion.
}

\paragraph{\textcolor{black}{Object-wise 3D Recovery.}}

Given the instance masks $\mathcal{M}$, we recover the local geometry and initial similarity transform of each segmented object using a single-view 3D reconstruction foundation model~\cite{chen2025sam}:
\begin{equation}
\mathcal{S}^{(0)}
=
\Phi_{\mathrm{rec}}(I,\mathcal{M})
=
\{(G_i,T_i^{(0)})\}_{i=1}^{N}.
\label{eq:sam3d_recovery}
\end{equation}
Here, $G_i$ denotes the local 3D geometry of the $i$-th object, and
$T_i^{(0)}=(R_i^{(0)},t_i^{(0)},s_i^{(0)})$
denotes its initial similarity transform, including rotation, translation, and scale.
This object-wise reconstruction provides the geometry and coarse placement of each relevant entity.
However, because individual objects are reconstructed independently, their estimated poses may produce spatial inconsistencies when assembled, such as floating objects or penetration into a support surface.

\paragraph{Support-aware Scene Optimization.}

Directly assembling independently recovered objects in a shared coordinate system can introduce spatial artifacts.
We therefore introduce a lightweight support-aware optimization to refine their placement.
In the tabletop interactions considered in this work, the dominant support surface provides a geometric reference for object placement.
Specifically, we estimate a support plane $\pi$ from the recovered geometry of the table or support object:
\begin{equation}
\pi
=
\left\{
x\in\mathbb{R}^{3}
\mid
n^{\top}x+d=0,\;
\|n\|_2=1
\right\}.
\label{eq:support_plane}
\end{equation}
Here, $n$ is the unit normal of the fitted support plane and $d$ is its scalar offset.
Since $n$ is normalized, $n^{\top}x+d$ gives the signed distance from a point $x$ to the support plane.

We then optimize the translation of each non-support object along the support normal $n$, while keeping its rotation and scale fixed:
\begin{equation}
\begin{aligned}
\min_{\{\delta_i\}}\quad
&\sum_{i\notin\mathcal{I}_{\mathrm{sup}}}
\left(
\mathcal{L}_{\mathrm{sup}}^{i}
+\lambda_{\mathrm{pen}}\mathcal{L}_{\mathrm{pen}}^{i}
+\lambda_{\mathrm{reg}}\delta_i^2
\right), \\
\text{with}\quad
&T_i=
\left(
R_i^{(0)},
t_i^{(0)}+\delta_i n,
s_i^{(0)}
\right).
\end{aligned}
\label{eq:support_optimization}
\end{equation}
Here, $\mathcal{I}_{\mathrm{sup}}$ denotes the set of support objects and $\delta_i$ is the scalar displacement of the $i$-th object along the support normal.
$\mathcal{L}_{\mathrm{sup}}^{i}$ encourages contact between the object bottom and the support surface,
$\mathcal{L}_{\mathrm{pen}}^{i}$ penalizes penetration into support structures,
and $\delta_i^2$ regularizes the displacement from the initial estimate.

After optimization, 
we obtain the final task-relevant scene representation as:
\begin{equation}
\mathcal{S}
=
\{(G_i,T_i)\}_{i=1}^{N}.
\label{eq:final_scene}
\end{equation}
\textcolor{black}{
The resulting representation contains the geometry and poses of the interaction object, target object, support structures, and task-relevant surrounding objects.
These quantities are subsequently used to determine the object start and goal states, detect collisions, replan object and hand trajectories when necessary, and condition hand synthesis.
}

\subsection{Scene-constrained Object Trajectory Planning}

Given the recovered scene $\mathcal{S}$ and the structured task specification
$\tau=(a,o_{\mathrm{int}},o_{\mathrm{goal}},r)$,
we plan a scene-constrained object trajectory that moves the interaction object from its recovered initial pose to a target pose determined by the goal region, spatial relation, and support geometry.
We first retrieve the initial 6DoF pose of $o_{\mathrm{int}}$ from $\mathcal{S}$ as the start point.
To estimate the target pose, we determine a candidate placement position from the location of $o_{\mathrm{goal}}$ and the spatial relation $r$, and adjust it using the corresponding support surface.
This adjustment reduces placement inconsistencies such as floating above or penetrating into the support surface.

\begin{equation}
\begin{aligned}
T_0^{\mathrm{obj}}
&=
\mathrm{ObjPose}(\mathcal{S},o_{\mathrm{int}}), \\
\hat{T}_1^{\mathrm{obj}}
&=
\mathrm{TargetPose}
(\mathcal{S},o_{\mathrm{int}},o_{\mathrm{goal}},r).
\end{aligned}
\label{eq:object_initial_target_pose}
\end{equation}
\textcolor{black}{
Here,
$T_0^{\mathrm{obj}}=(R_0,t_0)$
is the recovered initial pose of the interaction object, and
$\hat{T}_1^{\mathrm{obj}}=(\hat{R}_1,\hat{t}_1)$
is its support-adjusted target pose.
}

To generate a smooth motion between the initial and target poses, we construct a lifted cubic Bézier trajectory.
Let
$u_t=(t-1)/(T-1)$
be normalized time, and
$
\alpha_t
=
10u_t^3-15u_t^4+6u_t^5
$
be the minimum-jerk time parameter~\cite{flash1985coordination}.
Given a lift height $h$ and an upward direction $n_{\mathrm{up}}$ defined by the support normal or scene vertical direction, the Bézier control points are

$
p_0 = t_0, 
p_1 = t_0+h n_{\mathrm{up}}, 
p_2 = \hat{t}_1+h n_{\mathrm{up}}, 
p_3 = \hat{t}_1.
$

For object orientation, we use spherical linear interpolation to interpolate
between the initial rotation $R_0$ and the target rotation $\hat{R}_1$.
The resulting candidate object trajectory $T_t^{\mathrm{obj}}=(R_t^{\mathrm{obj}},t_t^{\mathrm{obj}})$ at frame $t$ is derived as:
\begin{equation}
\begin{aligned}
t_t^{\mathrm{obj}}
&=
\mathrm{Bezier}
(\alpha_t;p_0,p_1,p_2,p_3), \\
R_t^{\mathrm{obj}}
&=
\mathrm{SLERP}
(R_0,\hat{R}_1,\alpha_t).
\end{aligned}
\label{eq:object_trajectory}
\end{equation}

\textcolor{black}{
We then perform collision checking between the transformed interaction-object geometry along the candidate trajectory and the recovered geometry of surrounding scene objects.
If a collision is detected, the planner invokes a scene-aware replanning procedure to update the intermediate path while preserving the recovered initial pose and target pose.
Collision checking and replanning are repeated until the trajectory is collision-free with respect to the recovered scene geometry or the predefined replanning budget is reached.
}
The final object trajectory converts the parsed spatial goal into a continuous object-side motion and is subsequently coordinated with the synthesized hand motion.

\subsection{Hand Motion Synthesis}

Given the planned object trajectory, the remaining task is to synthesize the corresponding hand motion.
Directly optimizing high-dimensional hand parameters using only geometric objectives is under-constrained and may produce unstable finger configurations.
We therefore use contact as an intermediate representation:
a task-conditioned contact prior first predicts functional contact regions 
and a contact-conditioned hand-pose prior for grasp synthesis.

\paragraph{Task-conditioned Contact Prior.}
To guide hand synthesis for open-set object geometries, we first predict where the hand should contact the object surface.
\textcolor{black}{
Instead of learning category-specific contact patterns, the contact predictor should capture functional surface regions that are transferable across unseen object categories.
}
We therefore adopt a coarse-to-fine training strategy.
The contact predictor is first pre-trained on large-scale affordance data~\cite{yu2025seqafford,wang2026diffusionmodelsopenworldaffordance}, which provides coarse but generalizable supervision for functional object regions such as graspable parts and handles.
It is then fine-tuned using contact maps derived from HOI datasets, aligning the learned affordance prior with realistic hand-object contact distributions.
Inspired by~\cite{zhang2025openhoi}, we train a conditional VAE to model task-conditioned contact distributions on object surfaces.

Given the recovered interaction-object geometry $G$ and action type $a$, the contact prior predicts a point-wise contact probability map:
\begin{equation}
C
=
\Phi_{\mathrm{contact}}(G,a)
=
\{c_j\}_{j=1}^{M}.
\label{eq:contact_prior}
\end{equation}
Here, $M$ is the number of surface points sampled from $G$, and
$c_j\in[0,1]$
is the predicted contact probability of the $j$-th point.
The contact map serves two roles:
it conditions the grasp prior to initialize a contact-compatible grasp and serves as a contact constraint during optimization.

\paragraph{Contact-conditioned Hand Pose Prior.}

Given the predicted contact map $C$, we learn a contact-conditioned grasp prior that models plausible hand configurations around the object.
We represent the object-contact condition as
\begin{equation}
\mathcal{P}
=
\{(x_j,n_j,c_j)\}_{j=1}^{M},
\end{equation}
where $x_j$, $n_j$, and $c_j$ denote the position, surface normal, and predicted contact probability of the $j$-th object point, respectively.
A point-set encoder extracts the conditioning feature
$y=\Phi_{\mathrm{enc}}(\mathcal{P})$.
We then use a conditional variational autoencoder to model MANO~\cite{MANO} hand articulation conditioned on $y$.
The model contains a conditional prior encoder, a posterior encoder, and a decoder:
\begin{equation}
\begin{aligned}
p_{\phi}(z|y)
&=
\mathcal{N}
\big(
\mu_p(y),\sigma_p^2(y)
\big), \\
q_{\psi}(z|\theta,y)
&=
\mathcal{N}
\big(
\mu_q(\theta,y),\sigma_q^2(\theta,y)
\big), \\
\hat{\theta}
&=
D_{\omega}(z,y).
\end{aligned}
\label{eq:grasp_cvae}
\end{equation}
Here, $\theta$ denotes the MANO hand articulation parameters,
$p_{\phi}(z|y)$ predicts a latent distribution from the object-contact condition,
$q_{\psi}(z|\theta,y)$ approximates the posterior during training,
and $D_{\omega}$ decodes $z$ into hand articulation.

During training, the posterior and conditional prior are aligned using a KL-divergence loss, while the decoded hand is supervised using reconstruction losses on hand articulation and geometry.
At inference time, the posterior branch is discarded.
\textcolor{black}{
We sample or initialize $z$ from $p_{\phi}(z|y)$ and decode it into an initial hand articulation.
The global wrist rotation and translation are predicted by a separate wrist-pose head conditioned on the same object-contact feature $y$.
}

\paragraph{Contact-guided Latent Hand Optimization.}

Although the contact-conditioned prior provides a plausible initial grasp, the prediction may remain imperfect because of contact-estimation noise, reconstruction errors, and the domain gap between unseen objects and training data.
\textcolor{black}{
Direct optimization of high-dimensional MANO articulation parameters under these geometric cues can lead to unstable or implausible finger articulation.
}
Inspired by latent pose priors~\cite{SMPL-X:2019}, we instead refine the grasp in the learned latent space:
\begin{equation}
H(z,R^h,t^h)
=
\mathrm{MANO}
\left(
R^h,t^h,D_{\omega}(z,y)
\right),
\end{equation}
where $z$ controls finger articulation, and the global wrist rotation $R^h$ and translation $t^h$ determine the hand placement relative to the object.
We optimize the pose latent and wrist pose using
\begin{equation}
E
=
\lambda_{\mathrm{prior}}E_{\mathrm{prior}}
+
\lambda_{\mathrm{contact}}E_{\mathrm{contact}}
+
\lambda_{\mathrm{pen}}E_{\mathrm{pen}}.
\label{eq:latent_optimization_energy}
\end{equation}
Here,
$E_{\mathrm{prior}}$
regularizes $z$ toward the learned conditional prior,
$E_{\mathrm{contact}}$
attracts the hand surface toward the predicted high-probability contact regions,
and
$E_{\mathrm{pen}}$
penalizes hand-object interpenetration.
\textcolor{black}{
Optimizing through $z$ restricts finger articulation to configurations represented by the learned grasp prior, while the wrist variables align the hand with the reconstructed object.
}

After obtaining the optimized grasp,
\textcolor{black}{
we select an initial hand pose in the free space of the recovered task-relevant scene and first generate a candidate approach motion by smoothly interpolating the wrist pose and hand articulation toward the optimized grasp.
We perform collision checking between the candidate hand trajectory and the recovered non-interaction scene geometry.
If a collision is detected, the wrist trajectory is replanned while keeping the optimized terminal grasp fixed, and the hand articulation is interpolated along the updated approach trajectory.
}
During the object manipulation stage, we keep the optimized relative transformation between the hand and the object fixed, so that the hand shares the same global rigid motion as the manipulated object along the planned object trajectory while maintaining the optimized contact configuration.

\begin{figure*}[t]
    \centering
    \includegraphics[width=\textwidth]{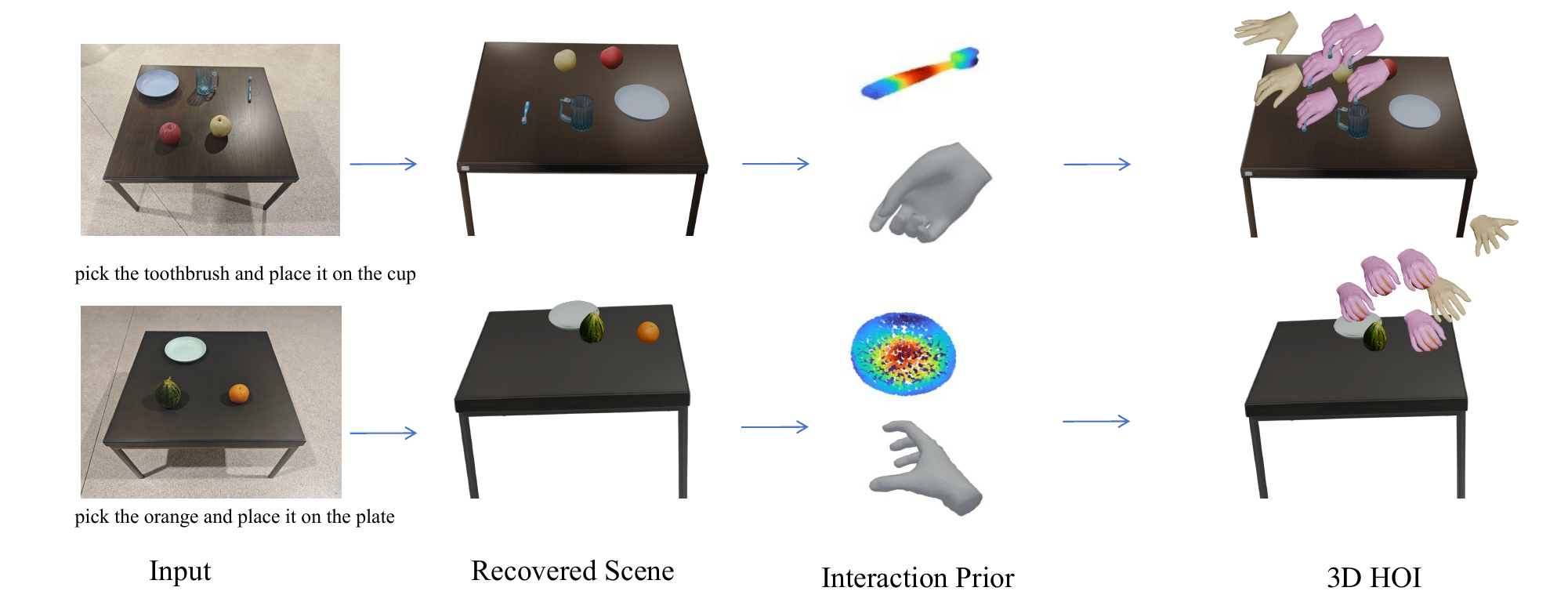}
    \caption{
    \textcolor{black}{
    Qualitative results on real-world photograph inputs.
    PhotoHOI recovers task-relevant geometry, plans the object trajectory, and synthesizes the corresponding hand motion.
    Yellow hands indicate the approach stage, while purple hands indicate the manipulation stage.
    }}
    \label{fig:visualization_results}
    \centering
    \includegraphics[width=\textwidth,keepaspectratio]{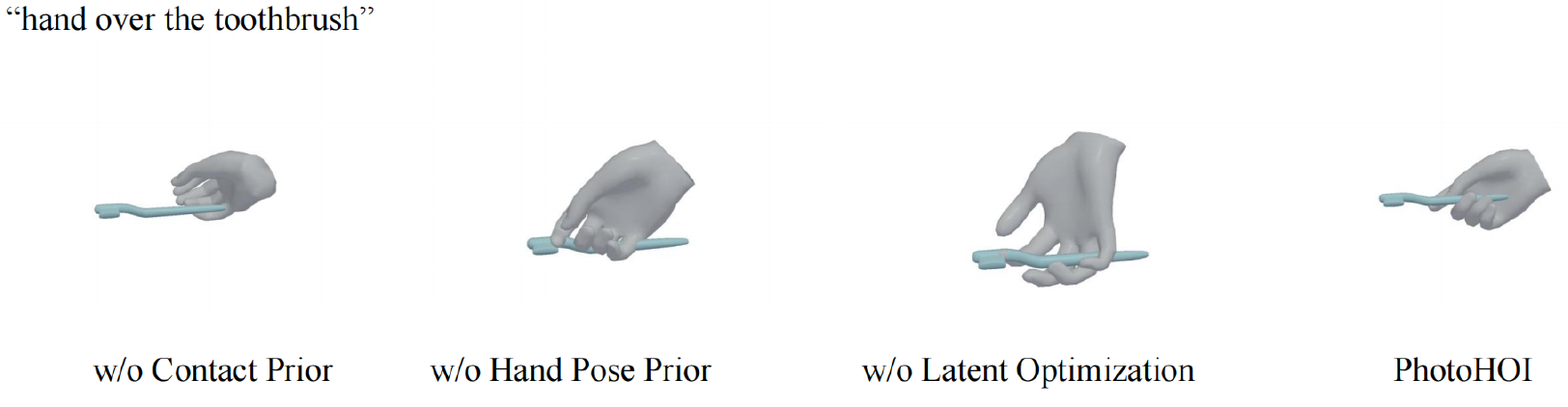}
    \caption{
    Qualitative ablation results of key components in PhotoHOI.
    }
  \label{fig:abl_vis}
\end{figure*}
\section{Experiments}
\textcolor{black}{
We evaluate PhotoHOI on standard HOI benchmarks and real-world photograph inputs.
Comparative experiments evaluate grasp quality and real-world task performance, while ablation studies examine the contributions of the contact prior, hand-pose prior, and latent optimization.
}

\subsection{Dataset}

\textcolor{black}{
The contact prior is first pre-trained on approximately 180K 3D affordance pairs~\cite{yu2025seqafford} and then aligned with contact maps derived from HOI data.
The hand-pose prior is trained on approximately 170K samples collected from GRAB~\cite{taheri2020grab}, H2O~\cite{kwon2021h2o}, ContactPose~\cite{brahmbhatt2020contactpose}, DexYCB~\cite{chao2021dexycb}, OakInk~\cite{yang2022oakink}, and HOGraspNet~\cite{cho2024dense}.
The full PhotoHOI model is trained on approximately 170K
samples collected from the aforementioned HOI datasets.
For comparisons with SOTA methods, we additionally train
benchmark-specific variants using only the training split of
the corresponding benchmark, i.e., GRAB for the GRAB
evaluation and H2O for the H2O evaluation.
The corresponding test splits are excluded from all training
stages.
}

\begin{table}[t]
\centering
{\small
\begin{tabular}{l|cccc}
\toprule
Method
& IV $\downarrow$
& ID $\downarrow$
& CR $\uparrow$
& SP $\downarrow$ \\
\midrule
GraspD
& 6.47 & 27.36 & 14.64 & 3.61 \\
DiffH2O
& 5.64 & 24.69 & 16.54 & 3.35 \\
Text2HOI
& 4.82 & 22.76 & 18.41 & 2.85 \\
OpenHOI
& 4.68 & 21.53 & 21.14 & 2.80 \\
\midrule
PhotoHOI (Ours)
& \textbf{3.83}
& \textbf{13.04}
& \textbf{29.43}
& \textbf{2.78} \\
\bottomrule
\end{tabular}
}
\caption{
Comparison with SOTA methods on GRAB.}
\label{tab:grab_results}
\end{table}
\begin{table}[t]
\centering
{\small
\begin{tabular}{l|cccc}
\toprule
Method
& IV $\downarrow$
& ID $\downarrow$
& CR $\uparrow$
& SP $\downarrow$ \\
\midrule
GraspD
& 9.65 & 31.27 & 17.63 & 3.79 \\
DiffH2O
& 8.45 & 28.43 & 18.47 & 3.35 \\
Text2HOI
& 7.88 & 25.29 & 20.63 & 2.83 \\
OpenHOI
& 6.34 & 22.37 & 21.05 & 2.81 \\
\midrule
PhotoHOI (Ours)
& \textbf{5.25}
& \textbf{20.89}
& \textbf{36.70}
& \textbf{2.79} \\
\bottomrule
\end{tabular}
}
\caption{
Comparison with SOTA methods on H2O.
}
\label{tab:h2o_results}
\end{table}

\subsection{Evaluation Metrics}

\textcolor{black}{
We evaluate the generated interactions from two perspectives:
grasp quality on standard HOI benchmarks and task-level performance on real-world photograph inputs.
}

\textbf{\textcolor{black}{Grasp Quality.}}
\textcolor{black}{
We report Interpenetration Volume (IV), Interpenetration Depth (ID), Contact Ratio (CR), and Self-Penetration (SP).
IV and ID measure hand-object penetration, CR measures contact coverage, and SP measures hand self-intersection.
Lower IV, ID, and SP and higher CR indicate better grasp quality.
}

\textbf{\textcolor{black}{Real-world Task Performance.}}
\textcolor{black}{
We report Task Success Rate (TSR) and Scene Consistency Rate (SCR).
TSR measures whether the instructed task is successfully completed, while SCR evaluates whether the recovered scene and generated interaction satisfy support, collision, and placement constraints.
}

\subsection{Main Results}

\paragraph{Comparison with SOTA Methods.}

\textcolor{black}{
We compare PhotoHOI with GraspD~\cite{turpin2022graspddifferentiablecontactrichgrasp}, DiffH2O~\cite{christen2024diffh2o}, Text2HOI~\cite{cha2024text2hoi}, and OpenHOI~\cite{zhang2025openhoi}.
As shown in Tabs.~\ref{tab:grab_results} and~\ref{tab:h2o_results}, PhotoHOI achieves better IV, ID, CR, and SP on both GRAB and H2O.
}

\begin{table}[t]
\centering
{\footnotesize
\setlength{\tabcolsep}{3.0pt}
\begin{tabular}{@{}l|cccccc@{}}
\toprule
Method
& IV $\downarrow$
& ID $\downarrow$
& CR $\uparrow$
& SP $\downarrow$
& TSR $\uparrow$
& SCR $\uparrow$ \\
\midrule
GraspD
& 22.46 & 106.35 & 5.85 & 3.02 & 24.95 & 22.70 \\
DiffH2O
& 21.95 & 98.32 & 5.49 & 2.92 & 29.60 & 25.30 \\
Text2HOI
& 20.67 & 96.97 & 6.94 & 2.83 & 42.15 & 38.20 \\
OpenHOI
& 19.25 & 78.41 & 10.71 & 2.76 & 43.90 & 39.35 \\
\midrule
PhotoHOI
& \textbf{10.32} & \textbf{49.21} & \textbf{11.24} & \textbf{1.89} & \textbf{63.05} & \textbf{62.75} \\
\bottomrule
\end{tabular}
}
\caption{
Real-world task evaluation on photograph inputs.
}
\label{real}
\end{table}

\paragraph{\textcolor{black}{Real-World Photograph Results.}}

\textcolor{black}{
We evaluate PhotoHOI across 30 real-world scenes containing
20 common objects. For each object, we conduct 100 randomized
trials and report the results averaged over all 2,000 trials.
Although all compared methods are provided with the same
task-relevant reconstructed scene geometry, they must still
synthesize an interaction that satisfies the instructed target
relation. As shown in Tab.~\ref{real} and
Fig.~\ref{fig:visualization_results}, PhotoHOI achieves higher
TSR and SCR, indicating more reliable task completion and
better scene consistency.
}

\subsection{Ablation Study}

\textcolor{black}{
We conduct ablation studies on GRAB using the same preprocessing, input conditions, and evaluation protocol.
Each variant removes one component from PhotoHOI: the contact prior, hand-pose prior, or latent optimization.
Results are reported in Tab.~\ref{ablation_study}, with visualizations provided in Fig.~\ref{fig:abl_vis}.
}

\textbf{Contact Prior.}
\textcolor{black}{
Removing the contact prior increases hand-object penetration and reduces CR.
This indicates that task-conditioned contact regions provide important geometric guidance for locating functional grasp areas.
}

\textbf{Hand-Pose Prior.}
\textcolor{black}{
Removing the hand-pose prior increases SP and slightly degrades the other grasp-quality metrics.
The learned pose distribution therefore provides a useful initialization and regularization for plausible hand articulation.
}

\textbf{Latent Optimization.}
\textcolor{black}{
Replacing latent optimization with direct MANO-parameter optimization produces slightly worse grasp-quality metrics.
Together with the qualitative results in Fig.~\ref{fig:abl_vis}, this suggests that latent-space refinement better preserves plausible hand articulation while satisfying contact and penetration constraints.
}

\begin{table}[t]
\centering
{\small
\setlength{\tabcolsep}{3.2pt}
\begin{tabular}{@{}l|cccc@{}}
\toprule
Method
& IV $\downarrow$
& ID $\downarrow$
& CR $\uparrow$
& SP $\downarrow$ \\
\midrule
w/o Contact Prior
& 5.23 & 16.86 & 28.95 & 2.81 \\
w/o Hand-Pose Prior
& 3.92 & 14.22 & 28.64 & 2.84 \\
w/o Latent Optimization
& 4.08 & 13.17 & 29.07 & 2.82 \\
\midrule
PhotoHOI
& \textbf{3.83}
& \textbf{13.04}
& \textbf{29.43}
& \textbf{2.78} \\
\bottomrule
\end{tabular}
}
\caption{
Ablation study on GRAB. We evaluate the effects of the contact prior, hand-pose prior, and latent optimization.
}
\label{ablation_study}
\end{table}
\section{Discussion}

\paragraph{Conclusion.}
PhotoHOI explores a more natural interface for 3D hand-object interaction synthesis, where the system starts from what users can easily provide: a real scene photograph and a language instruction, and produces a situated 3D interaction sequence. PhotoHOI grounds the task in a recovered 3D scene, plans the object-side motion, and synthesizes the corresponding hand motion with transferable contact and grasp priors. 
By reducing the need for manually prepared 3D object conditions, such as object scans and predefined trajectories, PhotoHOI makes HOI synthesis more accessible for practical content creation and embodied interaction.

\paragraph{Limitation and Future Work.}
PhotoHOI currently focuses on rigid-object interactions in tabletop scenes for 3D animation and content generation, and extending it to articulated or deformable objects would require modeling part-level affordances and object state changes. The framework also relies on single-image 3D recovery, where reconstruction errors may affect the generated results. 
In addition, our input setting is closely aligned with current embodied manipulation scenarios. Future work may explore retargeting the generated results to robotic embodiments.

\clearpage
\bibliography{main}

@String{Computer = "{IEEE} Computer" }

@String{Springer = "Springer-Verlag" }

@article{flash1985coordination,
  title={The coordination of arm movements: an experimentally confirmed mathematical model},
  author={Flash, Tamar and Hogan, Neville},
  journal={Journal of neuroscience},
  volume={5},
  number={7},
  pages={1688--1703},
  year={1985},
  publisher={Society for Neuroscience}
}

@article{dang2025svimo,
  title={Svimo: Synchronized diffusion for video and motion generation in hand-object interaction scenarios},
  author={Dang, Lingwei and Shao, Ruizhi and Zhang, Hongwen and Min, Wei and Liu, Yebin and Wu, Qingyao},
  journal={arXiv preprint arXiv:2506.02444},
  year={2025}
}

@article{zhang2026unihm,
  title={UniHM: Unified Dexterous Hand Manipulation with Vision Language Model},
  author={Zhang, Zhenhao and Liu, Jiaxin and Shi, Ye and Wang, Jingya},
  journal={arXiv preprint arXiv:2603.00732},
  year={2026}
}

@inproceedings{xu2023unidexgrasp,
  title={Unidexgrasp: Universal robotic dexterous grasping via learning diverse proposal generation and goal-conditioned policy},
  author={Xu, Yinzhen and Wan, Weikang and Zhang, Jialiang and Liu, Haoran and Shan, Zikang and Shen, Hao and Wang, Ruicheng and Geng, Haoran and Weng, Yijia and Chen, Jiayi and others},
  booktitle={Proceedings of the IEEE/CVF Conference on Computer Vision and Pattern Recognition},
  pages={4737--4746},
  year={2023}
}

@inproceedings{zhang2024graspxl,
  title={Graspxl: Generating grasping motions for diverse objects at scale},
  author={Zhang, Hui and Christen, Sammy and Fan, Zicong and Hilliges, Otmar and Song, Jie},
  booktitle={European Conference on Computer Vision},
  pages={386--403},
  year={2024},
  organization={Springer}
}

@inproceedings{zhong2025dexgrasp,
  title={Dexgrasp anything: Towards universal robotic dexterous grasping with physics awareness},
  author={Zhong, Yiming and Jiang, Qi and Yu, Jingyi and Ma, Yuexin},
  booktitle={Proceedings of the Computer Vision and Pattern Recognition Conference},
  pages={22584--22594},
  year={2025}
}

@inproceedings{li2024semgrasp,
  title={Semgrasp: Semantic grasp generation via language aligned discretization},
  author={Li, Kailin and Wang, Jingbo and Yang, Lixin and Lu, Cewu and Dai, Bo},
  booktitle={European Conference on Computer Vision},
  pages={109--127},
  year={2024},
  organization={Springer}
}

@inproceedings{cha2024text2hoi,
  title={Text2hoi: Text-guided 3d motion generation for hand-object interaction},
  author={Cha, Junuk and Kim, Jihyeon and Yoon, Jae Shin and Baek, Seungryul},
  booktitle={Proceedings of the IEEE/CVF Conference on Computer Vision and Pattern Recognition},
  pages={1577--1585},
  year={2024}
}

@inproceedings{christen2024diffh2o,
  title={Diffh2o: Diffusion-based synthesis of hand-object interactions from textual descriptions},
  author={Christen, Sammy and Hampali, Shreyas and Sener, Fadime and Remelli, Edoardo and Hodan, Tomas and Sauser, Eric and Ma, Shugao and Tekin, Bugra},
  booktitle={SIGGRAPH Asia 2024 Conference Papers},
  pages={1--11},
  year={2024}
}

@article{jiang2023motiongpt,
  title={Motiongpt: Human motion as a foreign language},
  author={Jiang, Biao and Chen, Xin and Liu, Wen and Yu, Jingyi and Yu, Gang and Chen, Tao},
  journal={Advances in Neural Information Processing Systems},
  volume={36},
  pages={20067--20079},
  year={2023}
}

@inproceedings{huang2025hoigpt,
  title={Hoigpt: Learning long-sequence hand-object interaction with language models},
  author={Huang, Mingzhen and Chu, Fu-Jen and Tekin, Bugra and Liang, Kevin J and Ma, Haoyu and Wang, Weiyao and Chen, Xingyu and Gleize, Pierre and Xue, Hongfei and Lyu, Siwei and others},
  booktitle={Proceedings of the Computer Vision and Pattern Recognition Conference},
  pages={7136--7146},
  year={2025}
}

@article{zhang2025openhoi,
  title={Openhoi: Open-world hand-object interaction synthesis with multimodal large language model},
  author={Zhang, Zhenhao and Shi, Ye and Yang, Lingxiao and Ni, Suting and Ye, Qi and Wang, Jingya},
  journal={arXiv preprint arXiv:2505.18947},
  year={2025}
}

@article{han2025touch,
  title={TOUCH: Text-guided Controllable Generation of Free-Form Hand-Object Interactions},
  author={Han, Guangyi and Zhai, Wei and Yang, Yuhang and Cao, Yang and Zha, Zheng-Jun},
  journal={arXiv preprint arXiv:2510.14874},
  year={2025}
}

@article{zhang2025manidext,
  title={Manidext: Hand-object manipulation synthesis via continuous correspondence embeddings and residual-guided diffusion},
  author={Zhang, Jiajun and Zhang, Yuxiang and An, Liang and Li, Mengcheng and Zhang, Hongwen and Hu, Zonghai and Liu, Yebin},
  journal={IEEE Transactions on Pattern Analysis and Machine Intelligence},
  year={2025},
  publisher={IEEE}
}

@article{zhang2026handx,
  title={HandX: Scaling Bimanual Motion and Interaction Generation},
  author={Zhang, Zimu and Zhang, Yucheng and Xu, Xiyan and Wang, Ziyin and Xu, Sirui and Zhou, Kai and Zhou, Bing and Guo, Chuan and Wang, Jian and Wang, Yu-Xiong and others},
  journal={arXiv preprint arXiv:2603.28766},
  year={2026}
}

@inproceedings{taheri2020grab,
  title={GRAB: A dataset of whole-body human grasping of objects},
  author={Taheri, Omid and Ghorbani, Nima and Black, Michael J and Tzionas, Dimitrios},
  booktitle={European conference on computer vision},
  pages={581--600},
  year={2020},
  organization={Springer}
}

@article{chen2025sam,
  title={Sam 3d: 3dfy anything in images},
  author={Chen, Xingyu and Chu, Fu-Jen and Gleize, Pierre and Liang, Kevin J and Sax, Alexander and Tang, Hao and Wang, Weiyao and Guo, Michelle and Hardin, Thibaut and Li, Xiang and others},
  journal={arXiv preprint arXiv:2511.16624},
  year={2025}
}

@inproceedings{jiang2024motionchain,
  title={Motionchain: Conversational motion controllers via multimodal prompts},
  author={Jiang, Biao and Chen, Xin and Zhang, Chi and Yin, Fukun and Li, Zhuoyuan and Yu, Gang and Fan, Jiayuan},
  booktitle={European Conference on Computer Vision},
  pages={54--74},
  year={2024},
  organization={Springer}
}

@article{deng2025human,
  title={Human-object interaction via automatically designed vlm-guided motion policy},
  author={Deng, Zekai and Shi, Ye and Ji, Kaiyang and Xu, Lan and Huang, Shaoli and Wang, Jingya},
  journal={arXiv preprint arXiv:2503.18349},
  year={2025}
}

@article{team2023gemini,
  title={Gemini: a family of highly capable multimodal models},
  author={Team, Gemini and Anil, Rohan and Borgeaud, Sebastian and Alayrac, Jean-Baptiste and Yu, Jiahui and Soricut, Radu and Schalkwyk, Johan and Dai, Andrew M and Hauth, Anja and Millican, Katie and others},
  journal={arXiv preprint arXiv:2312.11805},
  year={2023}
}

@article{ren2024grounded,
  title={Grounded sam: Assembling open-world models for diverse visual tasks},
  author={Ren, Tianhe and Liu, Shilong and Zeng, Ailing and Lin, Jing and Li, Kunchang and Cao, He and Chen, Jiayu and Huang, Xinyu and Chen, Yukang and Yan, Feng and others},
  journal={arXiv preprint arXiv:2401.14159},
  year={2024}
}

@inproceedings{yu2025seqafford,
  title={Seqafford: Sequential 3d affordance reasoning via multimodal large language model},
  author={Yu, Chunlin and Wang, Hanqing and Shi, Ye and Luo, Haoyang and Yang, Sibei and Yu, Jingyi and Wang, Jingya},
  booktitle={Proceedings of the IEEE/CVF Conference on Computer Vision and Pattern Recognition},
  pages={1691--1701},
  year={2025}
}

@inproceedings{kwon2021h2o,
  title={H2o: Two hands manipulating objects for first person interaction recognition},
  author={Kwon, Taein and Tekin, Bugra and St{\"u}hmer, Jan and Bogo, Federica and Pollefeys, Marc},
  booktitle={Proceedings of the IEEE/CVF international conference on computer vision},
  pages={10138--10148},
  year={2021}
}

@article{wang2022dexgraspnet,
  title={Dexgraspnet: A large-scale robotic dexterous grasp dataset for general objects based on simulation},
  author={Wang, Ruicheng and Zhang, Jialiang and Chen, Jiayi and Xu, Yinzhen and Li, Puhao and Liu, Tengyu and Wang, He},
  journal={arXiv preprint arXiv:2210.02697},
  year={2022}
}

@inproceedings{holl2018efficient,
  title={Efficient physics-based implementation for realistic hand-object interaction in virtual reality},
  author={H{\"o}ll, Markus and Oberweger, Markus and Arth, Clemens and Lepetit, Vincent},
  booktitle={2018 IEEE conference on virtual reality and 3D user interfaces (VR)},
  pages={175--182},
  year={2018},
  organization={IEEE}
}

@article{mangalam2024enhancing,
  title={Enhancing hand-object interactions in virtual reality for precision manual tasks},
  author={Mangalam, Madhur and Oruganti, Sanjay and Buckingham, Gavin and Borst, Christoph W},
  journal={Virtual Reality},
  volume={28},
  number={4},
  pages={166},
  year={2024},
  publisher={Springer}
}

@article{tang2023cafi,
  title={CAFI-AR: Contact-aware freehand interaction with ar objects},
  author={Tang, Xiao and Li, Ruihui and Fu, Chi-Wing},
  journal={Proceedings of the ACM on Interactive, Mobile, Wearable and Ubiquitous Technologies},
  volume={6},
  number={4},
  pages={1--23},
  year={2023},
  publisher={ACM New York, NY, USA}
}

@inproceedings{chen2025interactavatar,
  title={InteractAvatar: Modeling Hand-Face Interaction in Photorealistic Avatars with Deformable Gaussians},
  author={Chen, Kefan and Mohan, Sreyas and Theiss, Justin and Oprea, Sergiu and Sridhar, Srinath and Prakash, Aayush},
  booktitle={Proceedings of the IEEE/CVF International Conference on Computer Vision},
  pages={10410--10420},
  year={2025}
}

@inproceedings{xu2025interact,
  title={Interact: Advancing large-scale versatile 3d human-object interaction generation},
  author={Xu, Sirui and Li, Dongting and Zhang, Yucheng and Xu, Xiyan and Long, Qi and Wang, Ziyin and Lu, Yunzhi and Dong, Shuchang and Jiang, Hezi and Gupta, Akshat and others},
  booktitle={Proceedings of the Computer Vision and Pattern Recognition Conference},
  pages={7048--7060},
  year={2025}
}

@inproceedings{SMPL-X:2019,
  title = {Expressive Body Capture: 3D Hands, Face, and Body from a Single Image},
  author = {Pavlakos, Georgios and Choutas, Vasileios and Ghorbani, Nima and Bolkart, Timo and Osman, Ahmed A. A. and Tzionas, Dimitrios and Black, Michael J.},
  booktitle = {Proceedings IEEE Conf. on Computer Vision and Pattern Recognition (CVPR)},
  year = {2019}
}

@article{MANO,
      title = {Embodied Hands: Modeling and Capturing Hands and Bodies Together},
      author = {Romero, Javier and Tzionas, Dimitrios and Black, Michael J.},
      journal = {ACM Transactions on Graphics, (Proc. SIGGRAPH Asia)},
      volume = {36},
      number = {6},
      series = {245:1--245:17},
      month = nov,
      year = {2017},
      month_numeric = {11}
  }

@article{tu2026playerone,
  title={Playerone: Egocentric world simulator},
  author={Tu, Yuanpeng and Luo, Hao and Chen, Xi and Bai, Xiang and Wang, Fan and Zhao, Hengshuang},
  journal={Advances in Neural Information Processing Systems},
  volume={38},
  pages={145235--145261},
  year={2026}
}

@article{hao2026egosim,
  title={EgoSim: Egocentric World Simulator for Embodied Interaction Generation},
  author={Hao, Jinkun and Jia, Mingda and Wang, Ruiyan and Liu, Xihui and Yi, Ran and Ma, Lizhuang and Pang, Jiangmiao and Xu, Xudong},
  journal={arXiv preprint arXiv:2604.01001},
  year={2026}
}

@article{li2025scalable,
  title={Scalable vision-language-action model pretraining for robotic manipulation with real-life human activity videos},
  author={Li, Qixiu and Deng, Yu and Liang, Yaobo and Luo, Lin and Zhou, Lei and Yao, Chengtang and Zeng, Lingqi and Feng, Zhiyuan and Liang, Huizhi and Xu, Sicheng and others},
  journal={arXiv preprint arXiv:2510.21571},
  year={2025}
}

@article{luo2025being,
  title={Being-h0: vision-language-action pretraining from large-scale human videos},
  author={Luo, Hao and Feng, Yicheng and Zhang, Wanpeng and Zheng, Sipeng and Wang, Ye and Yuan, Haoqi and Liu, Jiazheng and Xu, Chaoyi and Jin, Qin and Lu, Zongqing},
  journal={arXiv preprint arXiv:2507.15597},
  year={2025}
}

@article{li2026gazevla,
  title={GazeVLA: Learning Human Intention for Robotic Manipulation},
  author={Li, Chengyang and Xiong, Kaiyi and Xu, Yuan and Qian, Lei and Wang, Yizhou and Zhu, Wentao},
  journal={arXiv preprint arXiv:2604.22615},
  year={2026}
}

@inproceedings{zhu2025evolvinggrasp,
  title={Evolvinggrasp: Evolutionary grasp generation via efficient preference alignment},
  author={Zhu, Yufei and Zhong, Yiming and Yang, Zemin and Cong, Peishan and Yu, Jingyi and Zhu, Xinge and Ma, Yuexin},
  booktitle={Proceedings of the IEEE/CVF International Conference on Computer Vision},
  pages={11665--11674},
  year={2025}
}

@inproceedings{liu2023contactgen,
  title={Contactgen: Generative contact modeling for grasp generation},
  author={Liu, Shaowei and Zhou, Yang and Yang, Jimei and Gupta, Saurabh and Wang, Shenlong},
  booktitle={Proceedings of the IEEE/CVF International Conference on Computer Vision},
  pages={20609--20620},
  year={2023}
}

@inproceedings{zhang2025bimart,
  title={Bimart: A unified approach for the synthesis of 3d bimanual interaction with articulated objects},
  author={Zhang, Wanyue and Dabral, Rishabh and Golyanik, Vladislav and Choutas, Vasileios and Alvarado, Eduardo and Beeler, Thabo and Habermann, Marc and Theobalt, Christian},
  booktitle={Proceedings of the Computer Vision and Pattern Recognition Conference},
  pages={27694--27705},
  year={2025}
}

@inproceedings{brahmbhatt2020contactpose,
  title={ContactPose: A dataset of grasps with object contact and hand pose},
  author={Brahmbhatt, Samarth and Tang, Chengcheng and Twigg, Christopher D and Kemp, Charles C and Hays, James},
  booktitle={European Conference on Computer Vision},
  pages={361--378},
  year={2020},
  organization={Springer}
}

@inproceedings{chao2021dexycb,
  title={Dexycb: A benchmark for capturing hand grasping of objects},
  author={Chao, Yu-Wei and Yang, Wei and Xiang, Yu and Molchanov, Pavlo and Handa, Ankur and Tremblay, Jonathan and Narang, Yashraj S and Van Wyk, Karl and Iqbal, Umar and Birchfield, Stan and others},
  booktitle={Proceedings of the IEEE/CVF conference on computer vision and pattern recognition},
  pages={9044--9053},
  year={2021}
}

@inproceedings{cho2024dense,
  title={Dense hand-object (ho) graspnet with full grasping taxonomy and dynamics},
  author={Cho, Woojin and Lee, Jihyun and Yi, Minjae and Kim, Minje and Woo, Taeyun and Kim, Donghwan and Ha, Taewook and Lee, Hyokeun and Ryu, Je-Hwan and Woo, Woontack and others},
  booktitle={European Conference on Computer Vision},
  pages={284--303},
  year={2024},
  organization={Springer}
}

@inproceedings{yang2022oakink,
  title={Oakink: A large-scale knowledge repository for understanding hand-object interaction},
  author={Yang, Lixin and Li, Kailin and Zhan, Xinyu and Wu, Fei and Xu, Anran and Liu, Liu and Lu, Cewu},
  booktitle={Proceedings of the IEEE/CVF conference on computer vision and pattern recognition},
  pages={20953--20962},
  year={2022}
}

@inproceedings{yan2025dexterous,
  title={Dexterous Manipulation Based on Prior Dexterous Grasp Pose Knowledge},
  author={Yan, Hengxu and Fang, Haoshu and Lu, Cewu},
  booktitle={2025 IEEE/RSJ International Conference on Intelligent Robots and Systems (IROS)},
  pages={1389--1396},
  year={2025},
  organization={IEEE}
}

@article{fang2025anydexgrasp,
  title={Anydexgrasp: General dexterous grasping for different hands with human-level learning efficiency},
  author={Fang, Hao-Shu and Yan, Hengxu and Tang, Zhenyu and Fang, Hongjie and Wang, Chenxi and Lu, Cewu},
  journal={arXiv preprint arXiv:2502.16420},
  year={2025}
}

@inproceedings{yu2025dynamic,
  title={Dynamic reconstruction of hand-object interaction with distributed force-aware contact representation},
  author={Yu, Zhenjun and Xu, Wenqiang and Xie, Pengfei and Li, Yutong and Anthony, Brian W and Zhang, Zhuorui and Lu, Cewu},
  booktitle={Proceedings of the IEEE/CVF International Conference on Computer Vision},
  pages={8590--8599},
  year={2025}
}

@article{chen2025fbi,
  title={FBI: Learning Dexterous In-hand Manipulation with Dynamic Visuotactile Shortcut Policy},
  author={Chen, Yijin and Xu, Wenqiang and Yu, Zhenjun and Tang, Tutian and Li, Yutong and Yao, Siqiong and Lu, Cewu},
  journal={arXiv preprint arXiv:2508.14441},
  year={2025}
}

@misc{turpin2022graspddifferentiablecontactrichgrasp,
      title={Grasp'D: Differentiable Contact-rich Grasp Synthesis for Multi-fingered Hands}, 
      author={Dylan Turpin and Liquan Wang and Eric Heiden and Yun-Chun Chen and Miles Macklin and Stavros Tsogkas and Sven Dickinson and Animesh Garg},
      year={2022},
      eprint={2208.12250},
      archivePrefix={arXiv},
      primaryClass={cs.RO},
      url={https://arxiv.org/abs/2208.12250}, 
}

@misc{dang2026harmohoiharmonizingappearance3d,
      title={HarmoHOI: Harmonizing Appearance and 3D Motion for Multi-view Hand-Object Interaction Synthesis}, 
      author={Lingwei Dang and Juntong Li and Zonghan Li and Hongwen Zhang and Liang An and Wei Min and Yebin Liu and Qingyao Wu},
      year={2026},
      eprint={2607.17097},
      archivePrefix={arXiv},
      primaryClass={cs.CV},
      url={https://arxiv.org/abs/2607.17097}, 
}

@misc{wang2026diffusionmodelsopenworldaffordance,
      title={Diffusion Models are Open-World Affordance Learners: Leveraging Generative Priors for 3D Affordance Learning}, 
      author={Hanqing Wang and Zhenhao Zhang and Kaiyang Ji and Mingyu Liu and Wenti Yin and yuchao chen and Zhirui Liu and Xiangyu Zeng and Tianxiang Gui and Hangxing Zhang and Jiahao Yuan and Zhiqing Cui and Jiaxin Liu and Zhiyuan Ma and Hui Xiong},
      year={2026},
      eprint={2508.01651},
      archivePrefix={arXiv},
      primaryClass={cs.CV},
      url={https://arxiv.org/abs/2508.01651}, 
}
\clearpage
\end{document}